\documentclass[lettersize,journal]{IEEEtran}

\usepackage[T1]{fontenc}    % use 8-bit T1 fonts
\usepackage{hyperref}       % hyperlinks
\usepackage{booktabs}       % professional-quality tables
\usepackage{nicefrac}       % compact symbols for 1/2, etc.
\usepackage{microtype}      % microtypography
\usepackage{xcolor}         % colors
\usepackage{epsfig} % for postscript graphics files
\usepackage{times} % assumes new font selection scheme installed

\usepackage{amsmath,amsfonts,amssymb}
\usepackage{float}
\usepackage{setspace}
\usepackage{multicol}
\usepackage{graphicx}
\usepackage{comment}
\usepackage{pgfplots}
\usepackage{tikz}
\usepackage{tikzscale}
\usepgfplotslibrary{groupplots}
\usepackage{svg}
\usepackage{siunitx}
\usepackage{cleveref}
\usepackage[font=small]{caption}
\usepackage[tight]{subfigure}
\usepackage{subfloat}
\usepackage{xr}
\hypersetup{
    urlcolor=black,
}
\usepackage[square,numbers]{natbib}
\usepackage[ruled,linesnumbered]{algorithm2e}
\usepackage{wrapfig}

\usepackage{amsmath}
\usepackage{amssymb}
\usepackage{graphicx}
\usepackage{textcomp}
\usepackage{xcolor}
\usepackage{pgfplots}
\usepackage{subfigure}
\usepgfplotslibrary{groupplots}
\usepackage{tikz}
\usepackage{tikzscale}
\usepackage{hyperref}
\usepackage{cleveref}
\usepackage{float}
\usepackage{multirow}
\usepackage{caption}
\usepackage{mathtools}
\usepackage{siunitx}
\usepackage{comment}
\usepackage{booktabs}
\usepackage{alphalph}
\usepackage{xspace}
\DeclareMathOperator*{\argmin}{\arg\!\min}

\newcommand{\ie}{\textit{i}.\textit{e}., }
\newcommand{\eg}{\textit{e}.\textit{g}., }
\newcommand{\st}{\text{s.t. }}

\newcommand{\myparagraph}[1]{\vspace{3pt} \noindent \textbf{#1} \ } % For cvpr/iccv submission

\crefname{chapter}{Chapter}{Chapter}
\Crefname{chapter}{Chapter}{Chapter}
\crefname{section}{Sec.}{Sec.}
\Crefname{section}{Sec.}{Sec.}
\crefname{figure}{Fig.}{Fig.}
\Crefname{figure}{Fig.}{Fig.}
\crefname{table}{Table}{Table}
\Crefname{table}{Table}{Table}
\crefname{equation}{Eq.}{Eq.}
\Crefname{equation}{Eq.}{Eq.}
\crefname{algocf}{Alg.}{Alg.}
\Crefname{algocf}{Alg.}{Alg.}

\newcommand{\LEGO}{LEGO}
\newcommand{\SLV}{StableText2Brick}
\newcommand{\LegoGPT}{\textsc{BrickGPT}}
\newcommand{\pp}{Prompt-to-Product}

\newcommand{\legostructure}{B=\{b_1, b_2, \dots, b_{N}\}\xspace}
\newcommand{\buildablespace}{\mathcal{B}_{\mathcal{I}, \mathcal{A}, \mathcal{S}}\xspace}
\newcommand{\alldesignspace}{\hat{\mathcal{B}}\xspace}
\newcommand{\ourspace}{\mathcal{B}(u \mid \mathcal{I}, \mathcal{A}, \mathcal{S})}
\newcommand{\designspace}{\mathcal{B}(u \mid \mathcal{I}, \mathcal{S})}
\newcommand{\genf}{f_{\LegoGPT{}}}
\newcommand{\buildf}{f_{\builder{}}}

\newcommand{\stabilityf}{f_\text{stability}}

\newcommand{\apex}{APEX-MR}
\newcommand{\builder}{\textsc{BrickMatic}}
\newcommand{\LEGOtm}{LEGO\texorpdfstring{\textsuperscript{\textregistered}}{}}

\SetKwInput{KwData}{Input}
\SetKwInput{KwResult}{Output}
\SetKwBlock{Loop}{loop}{end}
\SetKwComment{Comment}{/* }{ */}
\SetAlFnt{\small}
\SetAlCapFnt{\small}
\SetAlCapNameFnt{\small}
\SetAlCapHSkip{0pt}

\newcommand{\chapternote}[1]{{%
  \let\thempfn\relax% Remove footnote number printing mechanism
  \footnotetext[0]{\emph{#1}}% Print footnote text
}}

\begin{document}

\title{Prompt-to-Product: Generative Assembly via Bimanual Manipulation}

% \author{Anonymous Authors}
\author{Ruixuan Liu$^{*}$, Philip Huang$^{*}$, Ava Pun, Kangle Deng, Shobhit Aggarwal, Kevin Tang, Michelle Liu, \\
Deva Ramanan, Jun-Yan Zhu, Jiaoyang Li and Changliu Liu\\
Carnegie Mellon University \\
        % <-this % stops a space
\thanks{$^{*}$ Indicates equal contribution.}
}

% The paper headers
% \markboth{Journal of \LaTeX\ Class Files,~Vol.~14, No.~8, August~2021}%
% {Shell \MakeLowercase{\textit{et al.}}: A Sample Article Using IEEEtran.cls for IEEE Journals}

% \IEEEpubid{0000--0000/00\$00.00~\copyright~2021 IEEE}
% Remember, if you use this you must call \IEEEpubidadjcol in the second
% column for its text to clear the IEEEpubid mark.

\maketitle

\begin{abstract}

Creating assembly products demands significant manual effort and expert knowledge in 1) designing the assembly and 2) constructing the product.
This paper introduces \pp{}, an automated pipeline that generates real-world assembly products from natural language prompts.
Specifically, we leverage \LEGO{} bricks as the assembly platform and automate the process of creating brick assembly structures.
Given the user design requirements, \pp{} generates physically buildable brick designs, and then leverages a bimanual robotic system to construct the real assembly products, bringing user imaginations into the real world.
We conduct a comprehensive user study, and the results demonstrate that \pp{} significantly lowers the barrier and reduces manual effort in creating assembly products from imaginative ideas.
  
\end{abstract}

% \begin{IEEEkeywords}
% Article submission, IEEE, IEEEtran, journal, \LaTeX, paper, template, typesetting.
% \end{IEEEkeywords}

% A technical feature (regular or special issue) should meet the following requirements:

% No more than 9 magazine pages, so aim for no more than 4500 words of text
% No more than 10 equations
% No more than 20 references, unless it is a survey article
% No more than 10 figures
% Include at least one high quality color photograph of the robotic system
% Figures, tables, schematics, plots are very welcome
% PLEASE NOTE: figures need to be submitted in high-resolution, high-quality format such as JPEG, TIFF, EPS, etc. PDF files are not supported.

\section{Introduction}\label{sec:intro}

Transforming design concepts into real products is a complex, time-consuming process that requires both creativity and technical expertise.
Recent advances in generative artificial intelligence \cite{TripoSR2024,zhao2025hunyuan3d20scalingdiffusion} and additive manufacturing technologies \cite{URHAL2019335,bambu} have lowered the barrier to physical prototyping.
While existing techniques predominantly focus on rigid objects with fully connected, monolithic structures, they fall short when applied to assembly objects, \ie products composed of multiple interlocking components. 
Unlike monolithic bodies, assembly objects play a crucial role in the real world, as most engineered products, from toys and furniture to machines and electronics, are inherently modular and require assembly. 
These objects cannot be replaced by single rigid bodies without sacrificing practicality or functionality. 
Bringing 3D assembly designs into physical reality is particularly challenging, as assembly products must not only meet visual or aesthetic expectations, but also satisfy strict \textit{physical constraints}.

\begin{figure}
\centering
    \includegraphics[width=\linewidth]{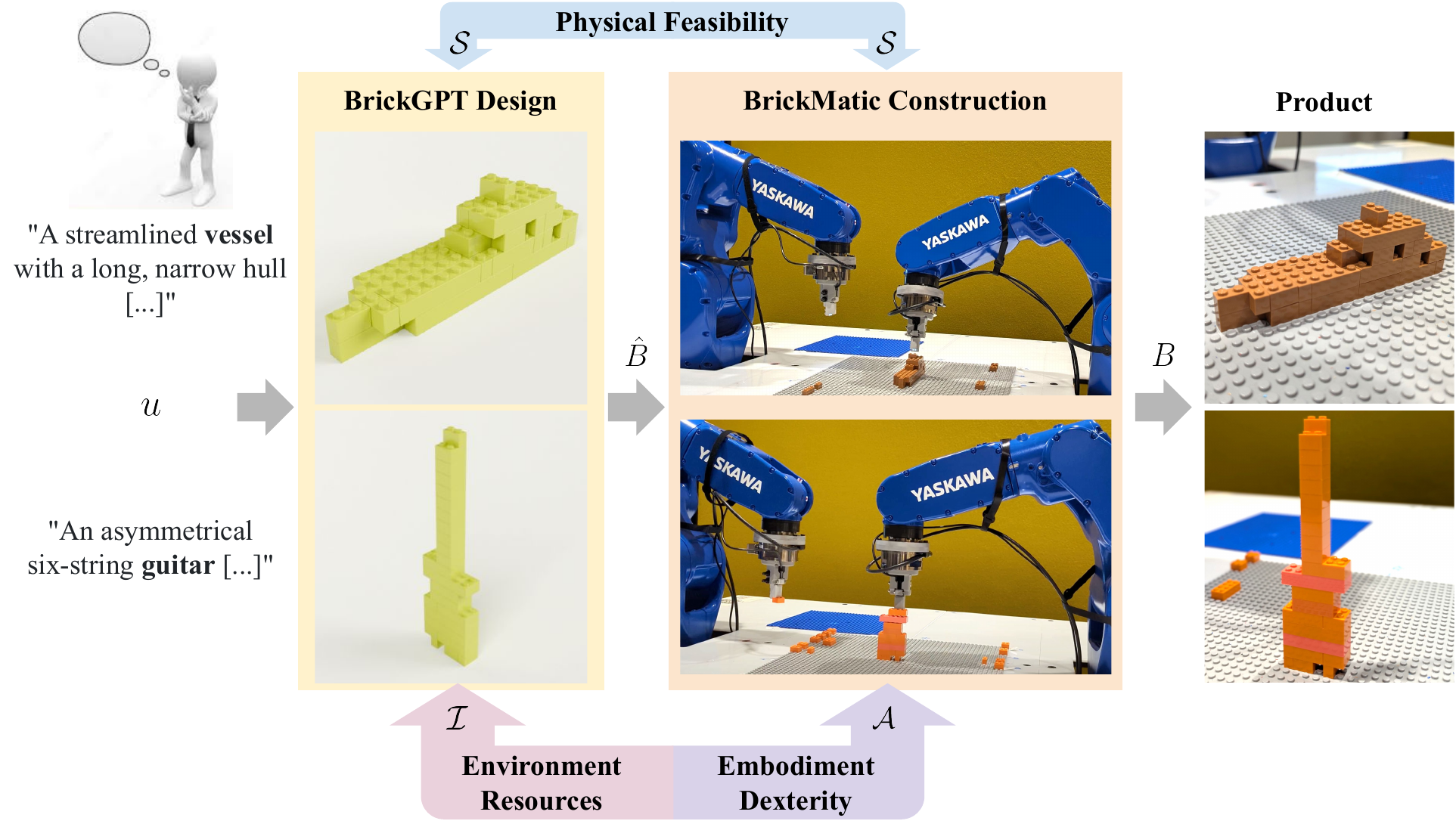}
    \caption{ \textbf{An Overview of \pp{}.} 
    Given the user design ideas, input as text prompts, \pp{} 1) generates the assembly designs using \LegoGPT{}, and 2) physically constructs the assembly products with \builder{}, bringing user imaginations into real products.
    \label{fig:intro-overview}}
    \vspace{-10pt}
\end{figure}

\myparagraph{Challenges.}
As illustrated in \cref{fig:intro-overview}, three important \textit{physical constraints} need to be satisfied.
First, the product needs to satisfy the \textit{environmental resource constraint}.
We must reason about the material available and build the assembly product only using the available inventory.
Second, the product should satisfy the \textit{embodiment dexterity constraint}.
It is important to understand the capability of the system and know what can be physically built within the skill set.
And most importantly, the assembly product should satisfy the \textit{physical feasibility constraint}.
It is critical to ensure that the assembly product is feasible as expected in the real world.
Due to the physical constraints mentioned above, it is challenging to bring assembly design ideas to life as real products.

\myparagraph{Related Works}
Recent work \cite{misra2025shapeshift} generates 2D puzzles using available building blocks but requires human effort to build them.
\citet{bang} transform a monolithic 3D object into an assembly by decomposing it into sub-components.
However, they assume an unlimited inventory and print \cite{bambu} the components, which later require manual assembly.
Our goal aligns closely with that of \citet{kyaw2024speech} and \citet{goldberg2025bloxnet}.
\citet{kyaw2024speech} generate an assembly design using prefabricated cuboctahedron blocks with magnet connections from user audio input, and build the assembly using a robot.
However, using customized cuboctahedron blocks as the assembly platform limits the replicability.
In addition, their system uses a simple rule-based check to assess physical stability and does not build collaboratively (\eg support), which prevents it from scaling to complex structures.
On the other hand, \citet{goldberg2025bloxnet} generate an assembly design from user text input and enable a robot to construct the structure physically using 3D printed blocks.
However, these blocks have smooth surfaces without interconnections, and thus have limited expressiveness.
Similarly, they use a single robot to construct the assembly object, which has very limited manipulation dexterity to build complex structures (\ie $> 10$ blocks).
In this work, our goal is to develop an integrated system that creates real-world 3D assemblies from user text input, allowing easy reproducibility, high customizability, and scalability up to complex structures with up to hundreds of components.
% The proposed \pp{} allows users to freely customize their design needs and facilitate a bimanual robotic system to collaboratively construct the assembly designs, which can scale up to complex structures with hundreds of components.

\myparagraph{Problem Statement.}
This paper studies the problem of translating natural language prompts into real-world 3D assembly products.
In particular, we leverage the \LEGOtm{} brick system as the assembly platform. 
LEGO bricks are low-cost and readily available, allowing easy replication for benchmarking.
In addition, they are inherently modular and highly reconfigurable, offering a vast design space through simple yet expressive interconnections. 
This makes them uniquely well-suited for this study.
By using LEGO bricks, this work aims to significantly reduce the manual effort required to turn user-generated text prompts into physical brick assembly structures—bridging the gap between imagination and realization. 
In the context of brick assembly, the physical constraints shown in \cref{fig:intro-overview} are defined as follows:
\begin{enumerate}
    \item Environment resources: we constrain the available inventory to contain 8 types of widely used bricks, including $1\times 1$, $1\times 2$, $1\times 4$, $1\times 6$, $1\times 8$, $2\times 2$, $2\times 4$, and $2\times 6$.
    \item Embodiment dexterity: we leverage general-purpose robotic arms with customized end-of-arm tools (EOAT) to build brick assemblies.
    \item  Physical feasibility: the brick assembly product should be physically stable and not collapse.
\end{enumerate}

\myparagraph{Prompt-to-Product.}
To this end, this paper introduces \pp{}, an automated pipeline that creates real-world brick assembly products from natural language prompts as illustrated in \cref{fig:intro-overview}. 
Due to the inherent complexity of assembly tasks, end-to-end methods \cite{kim2024openvla,black2024pi_0,intelligence2025pi_} often struggle with the long-horizon reasoning and fine-grained dexterity these tasks demand. 
\pp{} adopts a staged architecture to address the distinct yet interdependent challenges of translating natural language prompts into physical brick assemblies. 
This modular design separates the design generation and physical construction phases, allowing each to apply creative inference in the former and dexterous manipulation in the latter. 
Crucially, these modules are not isolated; they are tightly coupled through a shared physics reasoning module, which enforces global feasibility and maintains consistency across the pipeline.

In the first stage, \LegoGPT{} \cite{pun2025generating} generates brick assembly designs from user prompts, exploring the vast combinatorial space of brick structures while satisfying semantic and aesthetic user intents. 
In the second stage, \builder{} physically realizes the designs using a bimanual robotic system capable of dexterously executing long-horizon manipulation tasks.
At the core of \pp{} is a physics reasoning module, based on the structural stability analysis in \cite{Liu2024-go}, which plays a dual role. First, it constrains \LegoGPT{} to produce only physically buildable and stable designs given the available inventory. 
Second, it guides \builder{} through the physical assembly process by reasoning over intermediate stability and ensuring that the product remains constructible at every step.

As illustrated in \cref{fig:intro-overview}, this staged yet interconnected architecture enables \pp{} to effectively bridge user intent, structural design, and robotic execution—offering a coherent pipeline from prompt to product.
We conduct a comprehensive user study, including more than 20 participants.
% , to interact with either 1) \LegoGPT{}, 2) \builder{}, or 3) the full \pp{}.
The user study demonstrates that \pp{} significantly reduces the required manual effort and expert knowledge in creating brick assembly products from imaginative ideas.

\myparagraph{Contributions.}
Our contributions are listed as follows:
\begin{enumerate}
    \item We present \pp{} to create assembly products from users' design ideas in the form of text prompts.
    \item We introduce \builder{}, an integrated bimanual robotic system with enhanced dexterity capable of constructing customized brick assembly products.
    \item We conduct a comprehensive user study with participants from different backgrounds.
    The results demonstrate that \pp{} effectively reduces manual effort in creating assembly products from abstract ideas.
\end{enumerate}
% The rest of this paper is organized as follows: 
% \cref{sec:overview} discusses an overview of \pp{}.
% \Cref{sec:physics_reasoning} illustrates the physics reasoning module to evaluate the physical feasibility of brick assembly structures.
% \Cref{sec:brickgpt} details \LegoGPT{}, the module that generates assembly designs from user text prompts.
% \Cref{sec:brickmatic} introduces \builder{}, the bimnual robotic system that physically constructs brick structures given assembly designs.
% We discuss our user study in \cref{sec:experiment} as well as limitations and future works in \cref{sec:discussion}.

\section{Overview of \pp{}}
\label{sec:overview}

We formulate the \pp{} system as a staged process that maps a high-level user text prompt into a physically realized brick assembly through a sequence of modules. 
The staged design reflects the fundamentally different reasoning skills required at each step: creative semantic generation, physical feasibility evaluation, sequential assembly planning, and parallel robotic collaboration. 
Importantly, the stages are tightly coupled to ensure global consistency and feasibility.

\subsection{Problem Formulation}
We formulate the problem using the following notations:
\begin{itemize}
    \item $\{\cdot\}$: a set of elements where the order does not matter.
    \item $[\cdot]$: a sequence of elements where the order matters.
    \item $u \in \mathcal{U}$: a user prompt in natural language describing the design requirements of the assembly product.

    % Each node $v_r^i$ represents a motion configuration, manipulation, or perception skill for robot $i$. A type-1 edge $v_r^i \rightarrow v_{n+1}^i$ describes the execution order of a single robot. 
    % A type-2 edge $v_n^i \rightarrow v_{n'}^{i'}$ denotes a precedence order between two robots that constrains robot $n'$ to execute $v_{n'}^{i'}$ only after robot $n$ finishes $v_n^i$. 
        
    \item $B$: a brick structure layout represented as a set of bricks:
    \[
    \legostructure
    \]
    where $N$ is the number of bricks in the structure.
    Each brick \( b_i=\{c_i, p_i, \omega_i\} \) where $c_i$ denotes the brick type, \( p_i \in \mathbb{R}^3 \) is the brick pose in space, and $\omega_i$ is the planar orientation of the brick.

    \item \( \mathcal{I} \): the set of available inventory. 
    The environmental resource constraint enforces $\forall i\in\{1,2,\dots, N\}, c_i\in\mathcal{I}$.
    
    \item \( \mathcal{A} \): the set of skills that the system is capable of.
    The dexterity constraint requires the system to only use available skills $a_i\in \mathcal{A}$ to assemble $b_i$.

    \item $Q = [\{b^a_1, a_1\}, \{b^a_2, a_2\}, \dots, \{b^a_N, a_N\}]$: the sequence to build a brick structure $B$. 
    Each brick $b^a_i\in B$ and $a_i \in \mathcal{A}$ is the skill for assembling $b^a_i$.

    % \item $G$: system execution plan. 
    % Due to the facilitation of a bimanual system with two robots $r_1, r_2$ with $n_1, n_2$ degrees of freedom, the execution plan is represented as a temporal plan graph (TPG) \cite{tpg}, where $G = \{V, E\}$.
    % $V=[\{v_1^{r_1}, v_1^{r_2}\}, \{v_2^{r_1}, v_2^{r_2}\}, \dots, \{v_T^{r_1}, v_T^{r_2}\}]$ denotes the nodes in $G$, where $T$ is the total task horizon and represents trajectories for the bimanual system to execute the skill sequence $[a_1, a_2, \dots, a_N]$.
    % For $i, i'\in\{1, 2, \dots, T\}, j, j'\in\{1, 2\}$, each $v_i^{r_j}\in \mathbb{R}^{n_j}$ is a node in graph representing a robot configuration.
    % $E$ denotes the edges in $G$, which contains 1) type-1 edge $v_i^{r_j}\rightarrow v_{i+1}^{r_j}$ describing the execution order of a single robot, and 2) type-2 edge $v_i^{r_j}\rightarrow v_{i'}^{r_{j'}}$ representing a precedence order between robots that constrains robot $r_{j'}$ to execute $v_{i'}^{r_{j'}}$ only after robot $r_j$ finishes $v_i^{r_j}$.

    \item $G = \{V, E\}$: the bimanual system execution plan, represented as a temporal plan graph (TPG) \cite{tpg}. Each node in $V$ denotes a robot action to execute the skill sequence $[a_1, a_2, \dots, a_N]$, and each edge in $E$ indicates a precedence constraint between node executions.

    \item $S=[s_1;s_2;\dots;s_N]\in \mathbb{R}^N$: the stability evaluation of the entire structure $B$. Each $s_i\in \mathbb{R}$ is the stability score of the brick $b_i$. A brick is non-collapsing if $s_i > 0$.
    
    % \item $s_i\in \mathbb{R}$: the stability score of each brick $b_i$. A brick is non-collapsing in its layout if $s_i > 0$.

    % \item $S=[s_1;s_2;\dots;s_N]\in \mathbb{R}^N$: the stability evaluation of the entire structure $B$.
    
    \item \(\mathcal{S}=\{B \mid \forall i\in\{1, 2,\dots,N\}, s_i>0\}\): the set of all physically stable brick designs. The physical feasibility constraint requires $B\in \mathcal{S}$.
    
    % \item \( \mathcal{S} \): the set of allowable physical stability score $s_i\in\mathbb{R}$ of each brick $b_i$.
    % The physical feasibility constraint requires $\forall i, s_i>0$.

    \item $\buildablespace$: the space of all physically buildable designs with the given inventory and system (\ie structures that satisfy all physical constraints: $\mathcal{I}, \mathcal{A}, \mathcal{S}$).
    
    \item $\alldesignspace$: the set of all possible designs, either buildable or not.
    
    \item $ \alldesignspace(u) \subseteq \alldesignspace $: the set of all brick assembly designs that are semantically valid given the prompt \( u \).

\end{itemize}
Due to physical constraints, not all elements in \( \alldesignspace(u) \) are constructible. 
We define the feasible subset:
\[
 \ourspace \equiv \alldesignspace(u)\cap \buildablespace
\]
which includes only those assemblies that can be constructed with the given inventory and system capabilities.

The proposed \pp{} is a mapping from the user prompt $u$ to a physical assembly $B\in \ourspace$ that is buildable by the system with the given inventory. 
This paper focuses on constructing such a mapping efficiently, ensuring that the entire process, from prompt interpretation to physical realization, can be completed to support interactive and practical use cases. 
We also acknowledge that this is not the only way to instantiate such a mapping, but rather a structured and tractable approach enabled by modular reasoning.

\subsection{Physical Feasibility}
Reasoning physical feasibility is crucial in both designing and constructing the brick assembly.
It is important to ensure that the generated brick design is physically feasible in the first place; otherwise, it is impossible to build it.
Similarly, it is also critical to understand the physical feasibility during construction so that the system can reliably build the brick assembly rather than collapse it.
Evaluating the physical feasibility (stability) is non-trivial:
\begin{equation}
\label{eq:stability}
S=\stabilityf(B),
\end{equation}
where $B\in \alldesignspace$ can be any brick structure.
Conventional approaches \cite{tian2022assemble,goldberg2025bloxnet,zhang2025physics} rely on physics engines to simulate the structural stability. 
However, existing simulations fail to accurately model the deformations and interconnections between bricks, and thus, cannot correctly estimate the structural stability. 
\citet{pletzbrickfem} leverage the finite-element method to simulate high-fidelity behavior of brick structures, which is difficult to scale.
\citet{legolization} estimate the stability by analyzing the internal force distribution. 
However, their formulation only applies to structures that are connected as one piece, which is a subset of $\alldesignspace$.
We leverage the stability analysis in \cite{Liu2024-go} to model $\stabilityf(\cdot)$ and evaluate $S$ for any brick structure $B\in\alldesignspace$.
The details of $\stabilityf(\cdot)$ are discussed in \cref{sec:physics_reasoning}.

\subsection{Stage 1: Prompt-to-Design via \LegoGPT{}}
Given the input user prompt $u$, we define the generative mapping to generate the candidate brick layout design as:
\[
\hat{B} = f_{\text{design}}(u),
\]
where $\hat{B}=\{\hat{b}_1, \hat{b}_2,\dots,\hat{b}_N\}$.
The key requirement is that $\hat{B}$ should lie in $\ourspace$ in order to be physically built.
Since $\mathcal{A}$ is highly system dependent, 
% \eg a human builder would have a significantly larger $\mathcal{A}$ than a robot builder, 
we relax the constraint and require $\hat{B}\in \designspace$ in this stage, and consider $\mathcal{A}$ in the second stage.
By relaxing $\mathcal{A}$, the generated $\hat{B}$ becomes a human-buildable brick assembly design.
Conventional approaches decouple the constraints, \ie they leverage generative models \cite{TripoSR2024,zhao2025hunyuan3d20scalingdiffusion} to first generate the 3D shape that is visually and semantically appealing given $u$, and then use mesh-to-brick algorithms \cite{legolization,liu2024physics} to transform the 3D shape to a brick assembly structure that satisfy the physical constraints, \ie $\mathcal{I}$ and $\mathcal{S}$.
However, the generated 3D shape may have no solution for mesh-to-brick, meaning that it cannot be built using the brick inventory while being stable.
Hence, we leverage \LegoGPT{} \cite{pun2025generating}, an end-to-end approach, to generate the brick design $\hat{B}\in \designspace$ as
\begin{equation}
\label{eq:brickgpt_obj}
\begin{split}
    \hat{B} = & \genf(u),\\
    \st  \quad & c_i\in \mathcal{I}, ~\forall i\in\{1,2,\dots, N\},\\
    & \hat{B}\in \mathcal{S},
\end{split}
\end{equation}
where $N$ is internally determined by $\genf(\cdot)$.
The method for $\genf(\cdot)$ is discussed in \cref{sec:brickgpt}.

\begin{figure}
\centering
    \includegraphics[width=\linewidth]{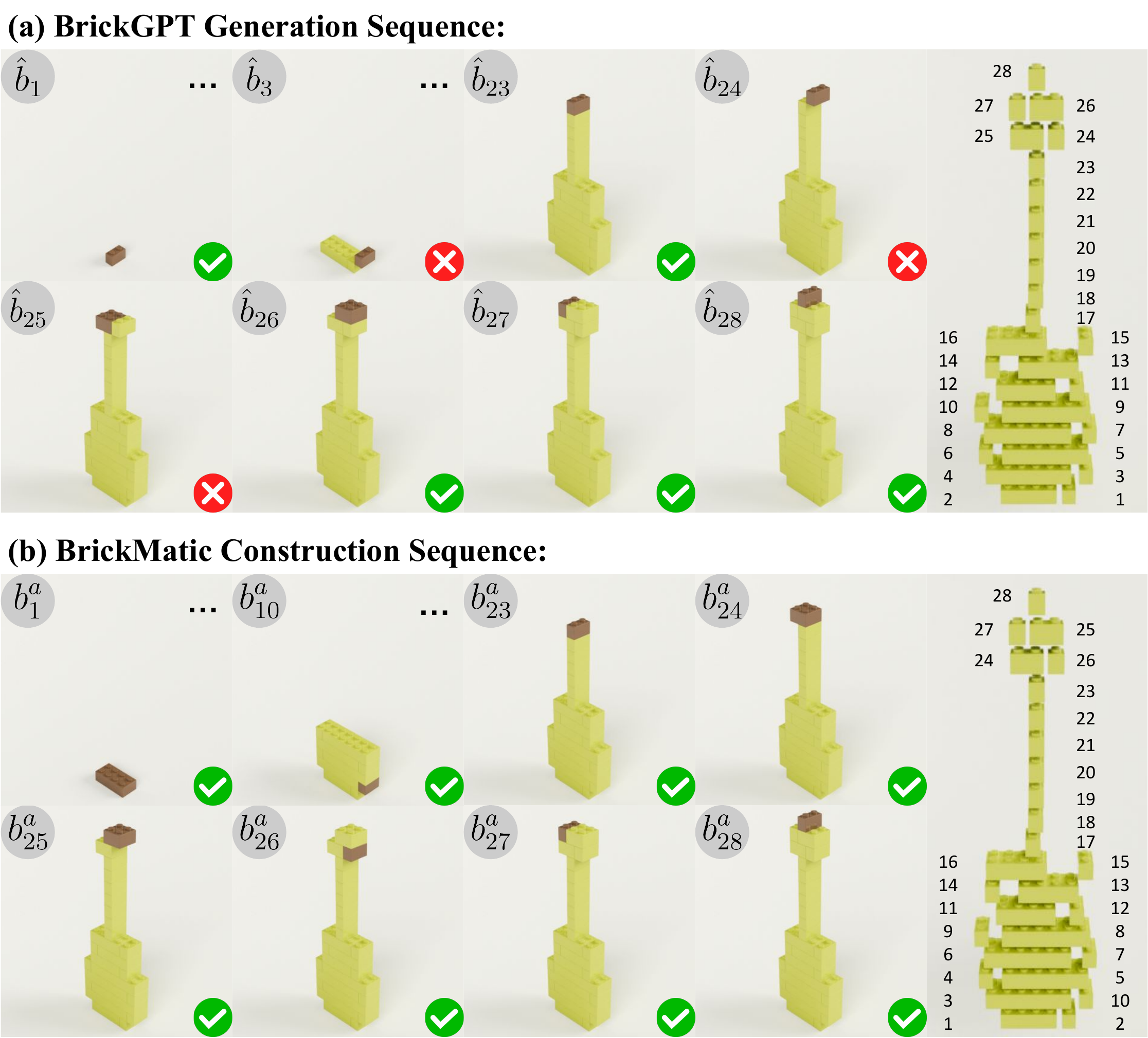}
    \caption{ \textbf{Brick Order.}
    (a) The sequence of \LegoGPT{} generating the design $\hat B$.
    (b) The assembly sequence $Q$ planned by \builder{} to physically construct the structure.
    The check mark indicates that the partial structure satisfies the physical feasibility constraint.
    \label{fig:sequence_diff}}
    \vspace{-10pt}
\end{figure}

\begin{figure*}
    \centering
    \includegraphics[width=\linewidth]{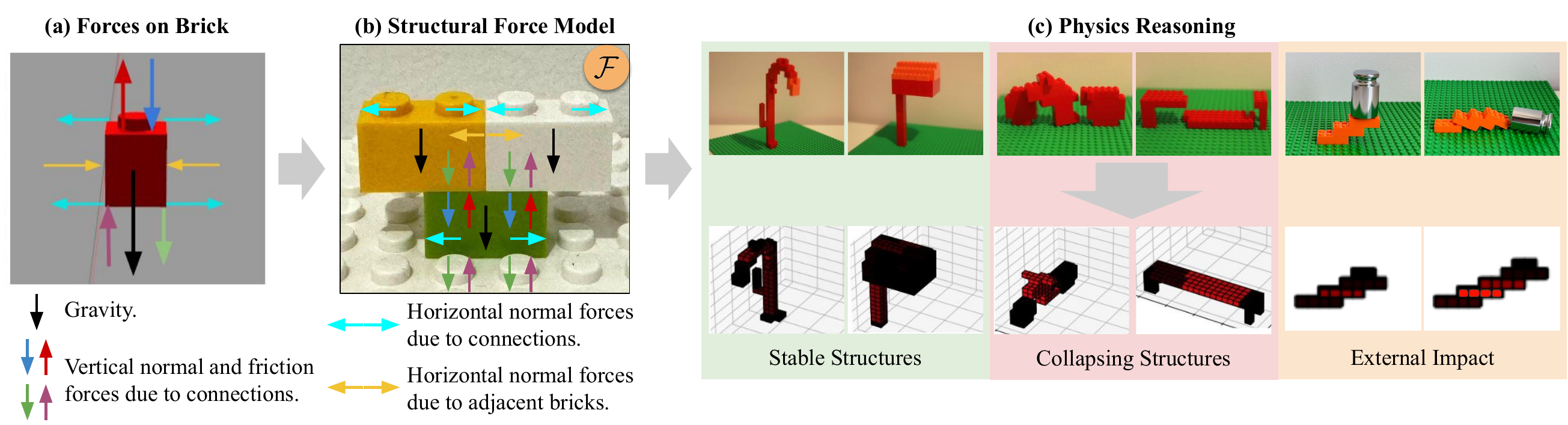}
    % \vspace{-10pt}
    \caption{\textbf{Assembly Physics Reasoning.} 
    (a) All possible forces that could be exerted on a brick. (b) The force model $\mathcal{F}$ of an example structure. (c) Given a brick structure $B$ (top), the physics reasoning solves $S=\stabilityf(B)$ to determine each brick's stability score $s_i$ and evaluate the structural stability (bottom).
    Darker color indicates higher stability. Red: collapsing bricks, \ie $s_i=0$.}
\label{fig:RAM_physics_reasoning}
    \vspace{-10pt}
\end{figure*}

\subsection{Stage 2: Design-to-Product via \builder{}}

Given the design \( \hat{B} \), the builder system constructs the brick assembly product as 
\[
B=f_{\text{build}}(\hat{B})
\]
At this step, the brick layout is not to be changed (essentially $B=\hat B$, though one is in the real physical space and the other is in the virtual space). But the building procedure needs to be carefully reasoned, taking the consideration of the physical constraints. 
Since the resource constraint is enforced in the first stage, we relax the constraint and only account for constraints $\mathcal{A}$ and $\mathcal{S}$.
Existing robotic systems often fall short with long-horizon reasoning and fine-grained dexterity, and thus, they only perform in simulation \cite{popov2017dataefficient,9341428} or build simple structures \cite{liu2023lightweight,goldberg2025bloxnet}.
Hence, we introduce \builder{}, a bimanual robotic system capable of building customized brick assemblies.
In particular, we expand $\mathcal{A}$ through embodiment design and skill learning discussed in \cref{sec:embodiment,sec:skillgraph}.
The physical construction is accomplished by a multi-level reasoning framework.
The first step is to reason over the assembly sequence $Q$, which enforces that each partial assembly $B^a_i=\{b^a_1, b^a_2, \dots, b^a_i\} \in \buildablespace$ is physically buildable given the system's capabilities.
As shown in \cref{fig:sequence_diff}, the planned $b^a_i\in Q$ can be different from $\hat{b}_i \in \hat{B}$ resulted from \LegoGPT{} in \cref{eq:brickgpt_obj}.
With the assembly sequence $Q$, \builder{} leverages \apex{} \cite{huang2025apexmr}, an asynchronous planning framework, to reason over 1) robot tasks (which robot performs which assembly step $\{b^a_i, a_i\}$), 2) robot motions (how to perform the assembly step without collision), and 3) collaborative executions (how to asynchronously collaborate to achieve better efficiency). 
Specifically, \builder{} generates a bimanual robot execution plan $G$ from $\hat{B}$ as:
\begin{equation}
\label{eq:brickmatic_obj}
\begin{split}
    G = & \buildf(\hat B)\\
    \st  \quad &B_i^a\in \mathcal{S},\\
    & a_i\in \mathcal{A}, ~\forall i\in\{1,2,\ldots, N\},\\ 
     % \quad & \gamma_j \text{ is collision free},\forall j\in\{1,2,\ldots, T\}.
\end{split}
\end{equation}
By executing $G$, \builder{} turns the design $\hat B$ into a physical assembly product $B$.
The details of \builder{} are discussed in \cref{sec:brickmatic}.

\subsection{\pp{} Pipeline Summary}

The entire \pp{} pipeline operates as follows:
\begin{align*}
\hat{B} &= \genf(u) \in \designspace \\
G &= \buildf(\hat B)\\
B &\leftarrow \mathbb{ROBOT\_EXECUTION}(G)\in \ourspace
\end{align*}
This staged but interconnected architecture ensures that high-level intent is preserved while satisfying the constraints of physical realizability and robotic execution. 
The central role of the physics reasoning module in constraining design generation and guiding robot construction provides a critical coupling between semantic generation and physical grounding.

\section{Assembly Physics Reasoning}\label{sec:physics_reasoning}

We leverage the stability analysis method in \cite{Liu2024-go} to estimate the physical feasibility, \ie structural stability $S$, of any brick structure $B\in\hat{\mathcal{B}}$.
Specifically, we optimize over force-balancing equations and solve for the force distribution to infer the stability of the brick structure.

\myparagraph{Structure Force Model.}
For a brick structure $B$, we reason the possible forces that could be exerted on it (\cref{fig:RAM_physics_reasoning}(a)), and construct the structural force model (\cref{fig:RAM_physics_reasoning}(b)), which consists of a set of forces $\mathcal{F}$, depending on the connections.
Each brick $b_i$ has $M_i$ forces exerting on it, where each force $F_i^j\in \mathcal{F}_i\subseteq \mathcal{F}, j\in\{1,2,\ldots, M_i\}$.
A structure is stable if all bricks can reach static equilibrium:
\begin{equation}\label{eq:static_equilibrium}
\begin{split}
    &\forall i\in\{1,2,\ldots, N\},\\ &\sum_j^{M_i}F_i^j=0, \quad \sum_j^{M_i}{\tau}_i^j \dot= \sum_j^{M_i}L_i^j\times F_i^j=0,
\end{split}
\end{equation}
where $L_i^j$ denotes the force lever corresponding to $F_i^j$.

\myparagraph{Structural Stability Analysis.}
To evaluate if a brick structure is physically stable, we solve its force distribution $\mathcal{F}$ by formulating the problem into a nonlinear program as
\begin{equation}\label{eq:force_obj}
\begin{split}
    \argmin_{\mathcal{F}}&\sum_i^{N}\Biggl\{|\sum_j^{M_i}F_i^j|+|\sum_j^{M_i}{\tau}_i^j|+\alpha \mathcal{D}_i^{\max}+\beta \sum \mathcal{D}_i\Biggl\},
    \end{split}
\end{equation}
subject to three constraints: 1) Non-negativity: $\forall F_i^j\in \mathcal{F}, F_i^j\geq 0$; 2) Non-coexistence: at any connecting point, the top brick cannot be supported while pulling by the bottom brick simultaneously; 3) Newton's third law: the forces exerted on adjacent bricks should be equal and opposite.
$\mathcal{D}_i\subset \mathcal{F}_i$ is the set of friction forces that drag $b_i$ down (\ie the green arrow in \cref{fig:RAM_physics_reasoning}(a)) and $\mathcal{D}_i^{\max}$ denotes the maximum value in $\mathcal{D}_i$.
Hyperparameters $\alpha, \beta$ are taking the same values as in \cite{Liu2024-go}.

\begin{figure*}
    \centering
    \includegraphics[width=\linewidth]{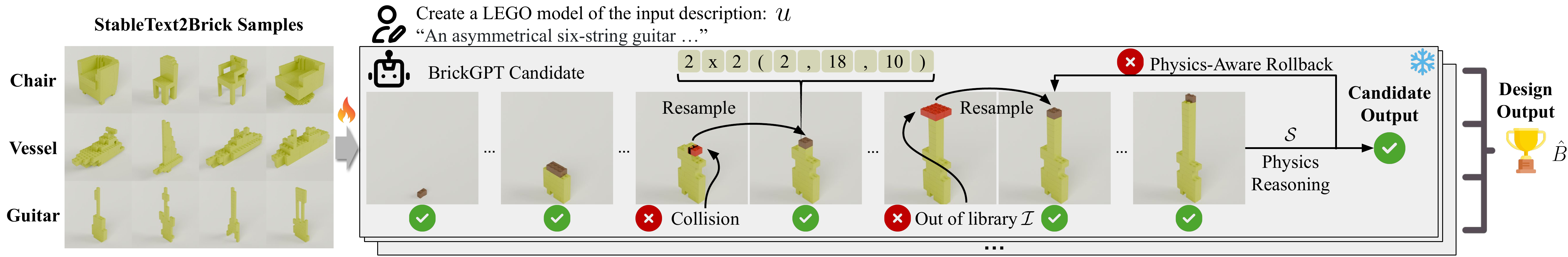}
    % \vspace{-20pt}
    \caption{\textbf{\LegoGPT{} for Generative Assembly Design.} \LegoGPT{} is fine-tuned on \SLV{} \cite{pun2025generating}. At inference time, multiple \LegoGPT{} candidates generate multiple assembly designs in parallel, and the best design is output as the final design.
    }
    \label{fig:brickgpt}
    \vspace{-10pt}
\end{figure*}

Solving the nonlinear program in \cref{eq:force_obj} finds a force distribution $\mathcal{F}$ that drives the structure to static equilibrium with the minimum internal friction.
% , suppressing the overall friction (\ie $\sum \mathcal{D}_i$) as well as avoiding extreme values (\ie $ \mathcal{D}_i^{\max}$).
From the solved $\mathcal{F}$, we obtain the structural stability $S$ and the score per brick as 
\begin{equation}\label{eq:stability_score}
    \begin{split}
        s_i=\begin{cases}
            0 & 
            \begin{array}{c}
                  \sum_j^{M_i}F_i^j \ne 0 \\ \vee \sum_j^{M_i}{\tau}_i^j \ne 0 \\
                  \vee \mathcal{D}_i^{\max}>F_T,
            \end{array}
            \\
            \frac{F_T - \mathcal{D}_i^{\max}}{F_T} & \text{otherwise},
        \end{cases}
    \end{split}
\end{equation}
where $F_T$ is a measured constant friction capacity between brick connections.
Higher $s_i$ indicates greater stability, while $s_i=0$ indicates an unstable brick that will cause structural failure: either the structure cannot reach static equilibrium (\ie $\sum_j^{M_i}F_i^j \ne 0 \vee \sum_j^{M_i}\tau_i^j \ne 0$) or the required friction exceeds the friction capacity of the material (\ie $\mathcal{D}_i^{\max}>F_T$). 
For a physically stable structure, we have $\forall i\in\{1, 2, \dots, N\}, s_i>0$. 
For any brick structure $B$, we can evaluate the structural stability $S=\stabilityf(B)$ by solving \cref{eq:force_obj,eq:stability_score}.

\Cref{fig:RAM_physics_reasoning}(c) illustrates examples of stable and collapsing brick structures with their visualized stability scores (\ie static stability). 
In addition, the stability analysis can account for external impact (\ie dynamic stability), determining whether the brick structure can withstand external weight.
The capability of numerically evaluating the structural stability mitigates the need for a reliable simulator. 
Consequently, \pp{} does not rely on a physics engine to understand real-world physics.
Instead, \pp{} leverages 1) the static stability to constrain \LegoGPT{} to generate physically buildable brick assembly designs $\hat{B}$ and 2) the dynamic stability to guide \builder{} to plan appropriate bimanual operations $G$ to construct the design.

\section{\LegoGPT{}: Generative Assembly Design}\label{sec:brickgpt}

% We leverage \LegoGPT{} \cite{pun2025generating} to generate physically buildable brick assembly structures from text prompts. 
% \LegoGPT{} is an end-to-end method that skips intermediate representations and directly generates the assembly design from a user text prompt as indicated in \cref{eq:brickgpt_obj}.
% It is built based on a large language model (LLM) backbone, \ie LLaMA-3 \cite{dubey2024llama}, to understand text prompt input.
% However, LLM generates text output instead of bricks.
% To ensure compatibility with LLM, a brick structure $\legostructure$ is represented in plain text, in which each line denotes a brick, as shown in \cref{fig:brickgpt}(a).
% A brick is denoted as ``\{$h$\}$\times$\{$w$\} (\{$x$\},\{$y$\},\{$z$\})'', where $h, w$ are brick dimensions in the $X$, $Y$ directions, and $x,y,z$ are its coordinates.
% Note that this representation is directly convertible to the representation in \cref{sec:overview}, in which $h$ and $w$ collectively determine the brick type $c$ and orientation $\omega$, and $x, y, z$ are equivalent to $p$.
% Also, this representation encodes the brick in a more geometrically explicit way, which is easier for model learning and independent of the representation of $\mathcal{I}$.
% As a result, \LegoGPT{} generates a brick assembly by outputting text responses word by word as illustrated in \cref{fig:brickgpt}(b).
% To generate physically buildable designs, \LegoGPT{} encodes physical constraints in both its training and inference phases.

We leverage \LegoGPT{} \cite{pun2025generating} to generate physically buildable brick assembly structures from text prompts. 
\LegoGPT{} is an end-to-end method that skips intermediate representations and directly generates the assembly design from a user text prompt as indicated in \cref{eq:brickgpt_obj}.
It is built based on a large language model (LLM) backbone to understand text prompt input.
However, LLM generates text output instead of bricks.
To ensure compatibility with LLM, \LegoGPT{} represents bricks in plain text as ``\{$h$\}$\times$\{$w$\} (\{$x$\},\{$y$\},\{$z$\})'', where $h, w$ are brick dimensions in the $X$, $Y$ directions, and $x,y,z$ are its coordinates.
As shown in \cref{fig:brickgpt}, a brick can be represented by 10 text tokens.
Note that this representation is directly convertible to the representation in \cref{sec:overview}, in which $h$ and $w$ collectively determine the brick type $c$ and orientation $\omega$, and $x, y, z$ are equivalent to $p$.
Also, this representation encodes the brick in a more geometrically explicit way, which is easier for model learning and independent of the representation of $\mathcal{I}$.
As a result, \LegoGPT{} generates a brick assembly by outputting text responses word by word.
To generate physically buildable designs, \LegoGPT{} encodes physical constraints in both its training and inference phases.

\subsection{Encoding Physical Constraints in Model Fine-tuning}

Despite having common knowledge, the LLM backbone lacks domain knowledge and the necessary geometric understanding for brick assembly.
Thus, \LegoGPT{} is fine-tuned on a large-scale brick assembly dataset, \ie \SLV{} \cite{pun2025generating}, as shown in \cref{fig:brickgpt}.
Each structure is paired with different captions.
Importantly, all brick structures in \SLV{} are verified to be physically buildable using the physics reasoning in \cref{eq:stability}, \ie $\text{\SLV{}}\subset \designspace$. 
\LegoGPT{} is fine-tuned using the stable text-brick pairs.

\subsection{Encoding Physical Constraints in Model Inference}

Fine-tuning on a buildable dataset enables \LegoGPT{} to gain knowledge in generating brick designs. 
However, it remains difficult to ensure the generated assembly design is always compliant with the physical constraints.
Thus, \LegoGPT{} further incorporates physical constraints into autoregressive inference. 
% Note that certain constraints (\eg structural stability) are more computationally demanding than other constraints (\eg resource).
Specifically, \LegoGPT{} decouples the physical constraints and hierarchically enforces them.

\myparagraph{Brick-by-Brick Rejection Sampling.}
During autoregressive inference, the constraints are relaxed to only consider resource and collision.
As depicted in \cref{fig:brickgpt}, after the model generates a brick $\hat b_i$, it immediately checks if $\hat b_i$ is valid, \ie 1) $\hat c_i\in \mathcal{I}$, and 2) $\hat b_i$ does not collide with the existing structure.
If $\hat b_i$ is invalid, \LegoGPT{} rejects it and resamples a new brick.

\myparagraph{Physics-Aware Rollback.}
Structural stability is applied at the end of the generation.
Specifically, the model uses the physics reasoning in \cref{eq:stability} to verify $\hat B\in \mathcal{S}$. 
Otherwise, \LegoGPT{} rolls back to a stable partial design, and regenerates from a partial design as shown in \cref{fig:brickgpt}.

\subsection{Multi-Head Generation}
\label{sec:brickgpt_multihead}
Due to the probabilistic nature, \LegoGPT{} could generate different designs with an identical text prompt input.
Thus, we present \LegoGPT{}++, which generates multiple designs in parallel to improve the design generation quality.
Given a prompt input $u$, we run eight \LegoGPT{} instances with different seeds and generate multiple candidate designs.
We render the outputs and select the best one according to the CLIP scores \cite{radford2021clip} as shown in \cref{fig:brickgpt}.

By integrating physical constraints, \ie resource and stability, in both model fine-tuning and inference, \LegoGPT{}++ generates brick assembly designs $\hat B$ that satisfy the users' customization requirements and are physically buildable in reality, \ie $\hat B\in \designspace$.

\section{\builder{}: Bimanual Assembly Construction}\label{sec:brickmatic}

We present \builder{}, a bimanual robotic system, to construct customized brick structures.
In particular, this work presents a robotic system design that enables general-purpose robot arms to robustly and precisely manipulate individual bricks.
In addition, a robot skill set $\mathcal{A}$ is introduced, which consists of a set of skill primitives, significantly expanding the system's dexterity.
To construct a full brick structure, we further develop a multi-level reasoning framework, which generates an execution plan $G$ as defined in \cref{eq:brickmatic_obj}.

\subsection{Robot Embodiment Design}
\label{sec:embodiment}

\begin{figure}
    \centering
    \includegraphics[width=\linewidth]{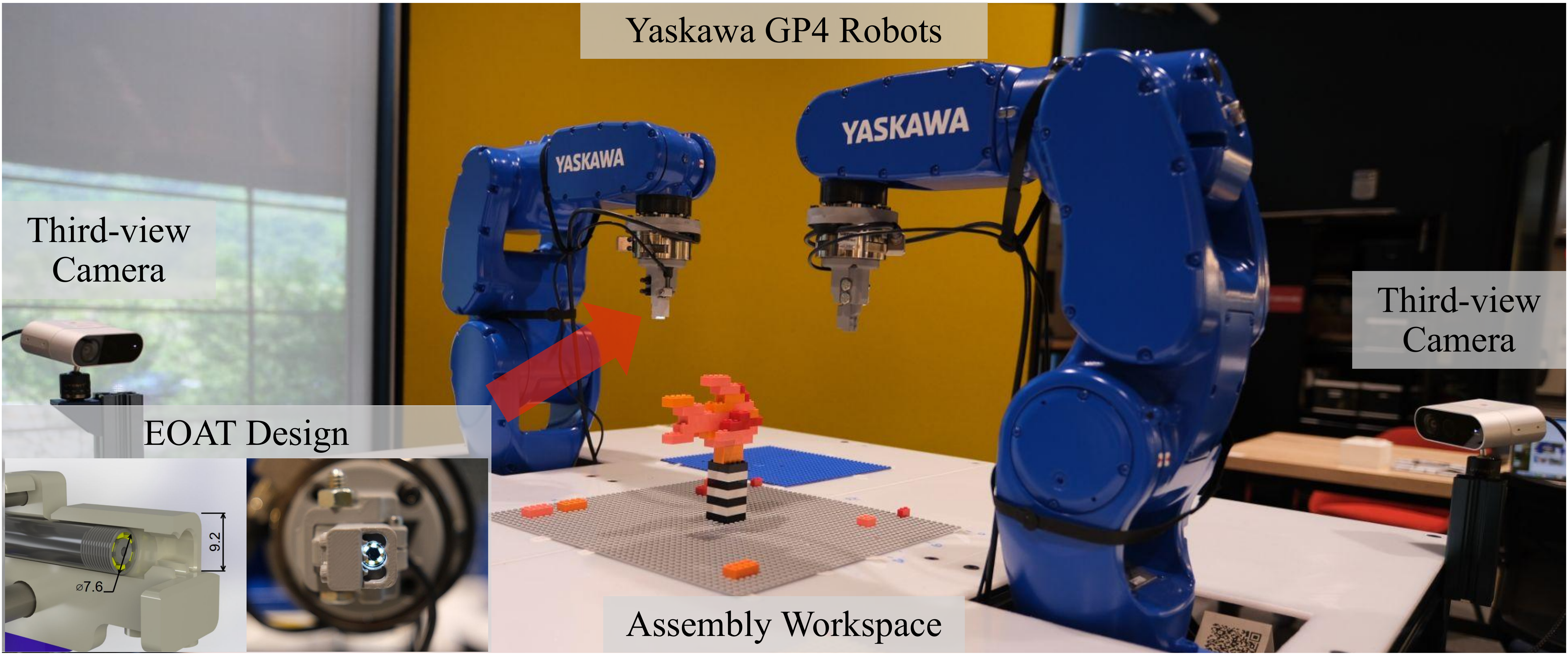}
    \caption{\textbf{\builder{} Embodiment Design.} 
    Our \builder{} includes two Yaskawa GP4 robots, each equipped with an FTS and an EiF EOAT, detailed in the bottom-left corner. 
    A baseplate is placed in the middle of the workstation for the robots to construct brick assembly structures.
    Two third-view cameras are placed diagonally.
    }
    \label{fig:embodiment}
    \vspace{-10pt}
\end{figure}

We present \builder{} as shown in \cref{fig:embodiment}.
\builder{} is a bimanual robotic system, consisting of two Yaskawa GP4 robot arms.
Each arm is equipped with an ATI force torque sensor (FTS) and a customized EOAT.
Two cameras are mounted diagonally at the workstation.
By facilitating dual arms, \builder{} covers a larger workspace, achieves higher task efficiency, and most importantly, is capable of collaboratively performing more dexterous manipulation tasks.

\builder{} leverages the Eye-in-Finger (EiF) \cite{tang2025eye} design as the EOAT to precisely and reliably manipulate bricks.
The design of EiF is detailed in the bottom left of \cref{fig:embodiment}.
The tooltip features a hollow interface that snaps over knobs on a brick and holds it from the top. 
It also has knobs on the side that allow it to securely hold a brick from the bottom.
Most importantly, EiF integrates an endoscope camera inside the tooltip. 
This type of camera is selected due to its low cost, small form factor, and built-in LED, which ensures consistent lighting. 
By integrating the camera into the EOAT, EiF enables close-proximity visual feedback without increasing the geometric volume. 

\subsection{Robot Skill Set}
\label{sec:skillgraph}

\begin{figure*}
\centering
    \includegraphics[width=\linewidth]{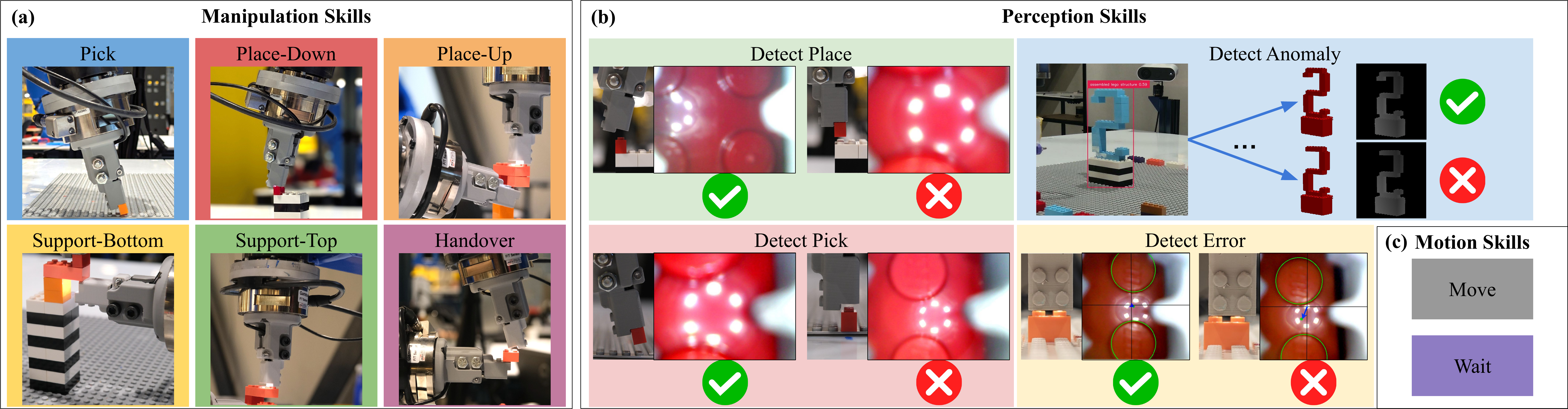}
    % \vspace{-15pt}
    \caption{\textbf{\builder{} Skill Set.} The skill set $\mathcal{A}$ of \builder{} with the presented embodiment. 
    \label{fig:skill}}
    \vspace{-10pt}
\end{figure*}

The multi-modality of \builder{} enables the system to perform a wide variety of different operations.
We parameterize the skill policies and construct a skill set $\mathcal{A}$ as shown in \cref{fig:skill}.
Our skill set has three major categories of skills: 1) manipulation skills, 2) perception skills, and 3) motion skills.

\myparagraph{Manipulation Skills.}
The manipulation skills represent structured, reusable action primitives that use sensor feedback (\eg force) to interact with the environment.
Following the design in \cite{huang2025apexmr}, our skill set includes six manipulation skills as shown in \cref{fig:skill}(a).

\begin{enumerate}
    \item \textbf{Pick}: The robot disassembles a brick from the existing structure and picks it up.
    Due to the EiF design, we leverage the insert-and-twist policy \cite{liu2023lightweight}, and the ``Pick'' skill is represented by two learnable parameters: a disassembly axis $O_d$ and a twisting angle $\theta_d$.
    
    \item \textbf{Place-Down}: The robot assembles a brick initially in its hand (holding from the top) and places it at a target location.
    Similarly, we use the insert-and-twist policy, and the action is represented by two learnable parameters: an assembly axis $O_a$ and a twisting angle $\theta_a$.
    
    \item \textbf{Place-Up}: The robot holds a brick from its bottom and assembles it in a location beneath another brick.
    We use the insert-and-twist policy, and the skill is represented by two learnable parameters: an assembly axis $O^u_a$ and a twisting angle $\theta^u_a$.
    
    \item \textbf{Support-Bottom}: The robot supports the brick structure from the bottom.

    \item \textbf{Support-Top}: The robot supports the brick structure from the top.

    \item \textbf{Handover}: One robot is holding a brick from the top and places it onto the EiF EOAT of the other robot.
    After the operation, the other robot holds the brick from its bottom.
    This skill shares the same $O_a$ and $\theta_a$ from the ``Place-Down'' skill.
\end{enumerate}
All manipulation skills are goal-conditioned, \ie conditioned on the goal location to support or pick/place a brick, and use FTS feedback to 1) avoid excessive pressure when picking and placing bricks, and 2) ensure a slight contact when supporting to prevent breaking the structure.
We follow the learning technique in \cite{liu2023lightweight} to optimize the twisting axes $O_d, O_a, O_a^u$ and twisting angles $\theta_d, \theta_a, \theta_a^u$.

\myparagraph{Perception Skills.}
\builder{} can robustly manipulate individual bricks with the manipulation skills.
However, contingencies could occur when building full structures.
For instance, bricks assembled earlier could be loosened due to later operations, causing the structure to tilt, or even collapse.
Thus, we introduce new perception skills, which leverage either the EiF in-hand camera or the third-view camera to update the \builder{}'s knowledge on the environment and detect if any contingency occurs while the robots remain in place.
We present four perception skills as illustrated in \cref{fig:skill}(b).

\begin{enumerate}
    \item \textbf{Detect Place}: 
    The system detects if a place operation (\ie ``Place-Down'' or ``Handover'') is successful using the EiF camera.
    Essentially, this skill detects if the brick is released after the operation.
    Specifically, we learn a binary classifier on top of the DINOv2 features \cite{oquab2023dinov2} of the EiF camera input.
    The classifier determines that the operation is successful if the brick is released.
    The ``Detect Place'' block in \cref{fig:skill}(b) depicts examples of camera views and their corresponding output.
    
    \item \textbf{Detect Pick}:
    The system detects if a ``Pick'' skill is successfully performed using the EiF camera.
    The same classifier determines that the operation is successful if the brick is in-hand.
    The ``Detect Pick'' block in \cref{fig:skill}(b) depicts examples of detections.

    \item \textbf{Detect Anomaly}:
    Due to the non-rigid connections between bricks, previously established connections could be loosened due to later operations, causing the structure to tilt or even collapse.
    To address this, we use third-view cameras as shown in \cref{fig:embodiment} to detect anomaly, \ie structural failure. 
    As shown in the ``Detect Anomaly'' block in \cref{fig:skill}(b), we compare the real camera view of the structure (\ie the bounding box) with the desired camera view rendered in simulation \cite{1389727} (\ie the red visuals and their depth maps). 
    If the real observation does not match the desired simulation rendering, then we classify the current scene as an anomaly.
    
    \item \textbf{Detect Error}:
    The system's knowledge of the environment could deviate from reality due to, for instance, calibration error, tilted structures, etc. 
    Following \cite{tang2025eye}, we use the endoscopic camera to detect and correct model mismatches. 
    As shown in the ``Detect Error'' block in \cref{fig:skill}(b), we use a fine-tuned YOLOv8-seg \cite{yolov8} to detect brick knobs and calculate the offset between the actual brick position and the center of EiF. 
    The robot adjusts its pose accordingly to ensure accurate alignment with the brick, which is essential for subsequent manipulation skills, \eg ``Pick'', ``Place-Down'', etc.
\end{enumerate}

\myparagraph{Motion Skills.}
The motion skills are skills that do not physically interact with the environment.
Our skill set includes two fundamental motion skills: ``Wait'' and ``Move'' shown in \cref{fig:skill}(c). 
The ``Move'' skill is responsible for collision-free movements between waypoints using motion planners, \eg \cite{RRT-Connect,jpc}.
The ``Wait'' skill pauses the robot's motion.

\subsection{Multi-Level Reasoning}
\label{sec:mr_reasoning}

\begin{figure*}
    \centering
    \includegraphics[width=\linewidth]{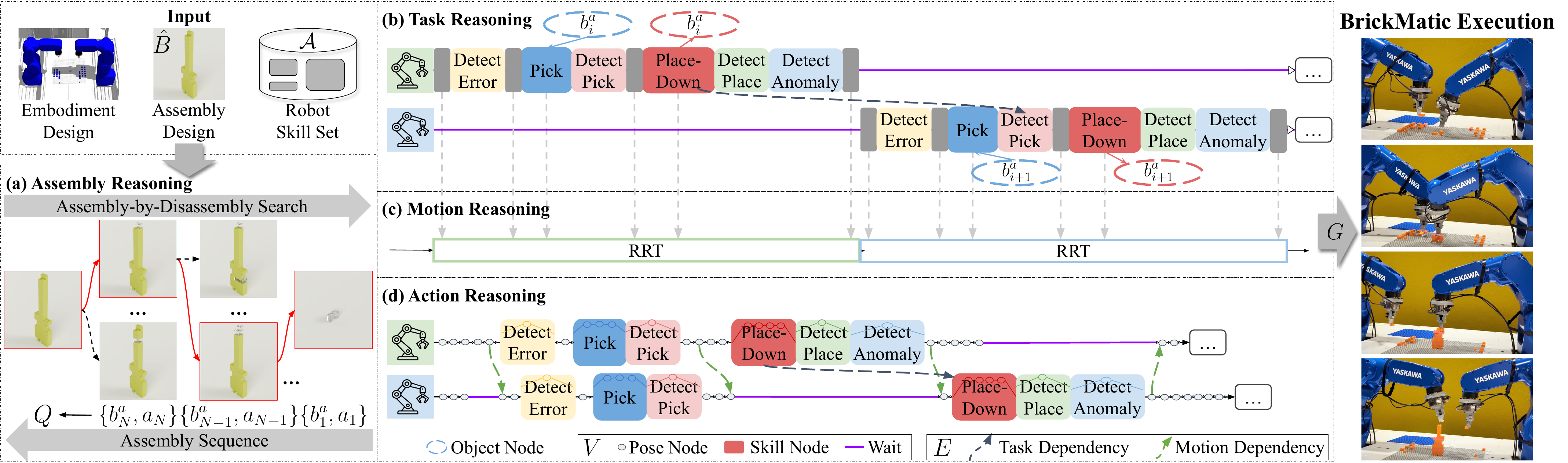}
    \caption{\textbf{\builder{} Multi-Level Reasoning for Bimanual Assembly Construction.} 
    Given an assembly design, \builder{} (a) reasons the assembly structure and generates a physically executable step-by-step assembly task plan $Q$.
    It then (b) reasons the tasks, distributes the assembly steps to different robots, and builds a sequential task plan using the available skills from the skill set $\mathcal{A}$.
    After that, \builder{} does (c) motion reasoning to plan the motions for each robot to perform the tasks, and (d) action reasoning to convert the sequential motion plan into a parallel, asynchronous action plan, \ie a TPG $G=\{V, E\}$, for physical execution. The colors of the skill nodes correspond to the colors in the skill set in \cref{fig:skill}.
    }
    \label{fig:multilevel_reasoning}
    \vspace{-10pt}
\end{figure*}

The embodiment design and the skill set provide \builder{} enhanced dexterity to reliably manipulate individual bricks.
To construct a complete brick structure, we present a multi-level reasoning framework to plan and coordinate the bimanual system.
As illustrated in \cref{fig:multilevel_reasoning}, the presented multi-level reasoning framework considers the embodiment design, the skill set $\mathcal{A}$, and the brick design $\hat B$, and generates an execution plan $G$ that controls the bimanual system to construct the physical brick assembly structure $B\in\ourspace$.

\myparagraph{Assembly Reasoning.}
Even though \LegoGPT{} generates the brick design brick-by-brick, it only ensures the final design $\hat B$ is physically buildable.
Hence, intermediate structures could be unstable as illustrated in \cref{fig:sequence_diff}(a). 
Thus, given the brick assembly design $\hat B$, the first step is to plan a physically executable assembly sequence by reordering the bricks as depicted in \cref{fig:sequence_diff}(b).
Specifically, we use assembly-by-disassembly search \cite{tian2022assemble} to search for an assembly sequence.
As illustrated in \cref{fig:multilevel_reasoning}(a), we start from the goal state, \ie the final assembly design, and disassemble a brick step by step until there is no brick remaining.
The assembly sequence is obtained by reversing the disassembly sequence.

However, for each disassembly step, multiple bricks could be disassembled, and it is crucial to choose the physically executable one so that the planned assembly sequence is executable by \builder{}.
To this end, we leverage the action mask presented in \cite{liu2024physics} to evaluate if a brick $\hat b_i$ is removable from a partial structure $\hat B_i$ using skill $a_i\in\mathcal{A}$ by assessing the following properties:
\begin{enumerate}
    \item \textbf{Operability}: there is space for the system to perform the skill $a_i$.
    Let $\mathcal{V}_{a_i}$ be the volume occupied by $a_i$, we ensure $\mathcal{V}_i\cap \mathcal{V}_{a_i}=\varnothing$, where $\mathcal{V}_i$ denotes the volume occupied by $\hat B_i$.
    Since $a_i$ is a goal-conditioned skill primitive, we can easily estimate the occupied space.
    
    \item \textbf{Static stability}: the structure after removing the brick remains physically stable. 
    Let $\hat B'_i$ be the structure resulted from removing $\hat b_i$ from $\hat{B}_i$, we ensure $\hat B'_i\in \mathcal{S}$. 
    
    \item \textbf{Dynamic stability}: the structure $\hat B_i$ is stable under the impact of $a_i$.
    Since the physics reasoning $\stabilityf(\cdot)$ can account for external impact by adding virtual bricks as shown in \cref{fig:RAM_physics_reasoning}, we model the robot place operation as a virtual brick $\hat b_p$ with heavy mass, \ie 1kg. 
    If support is needed, we add an additional virtual brick $\hat b_s$ with mass being $-1$kg. 
    We apply the physics reasoning to the virtual structure $\hat B_i^+=\hat B_i\cup\{\hat b_p, \hat b_s\}$ and ensure $\hat B_i^+ \in \mathcal{S}$.
\end{enumerate}
By ensuring the above criteria, we ensure the disassembly step $\{\hat b_i, a_i\}$ is physically executable. 
Note that $a_i$ can be a place skill alone or in combination with a support skill.
We use the action mask to prune any invalid disassembly steps, which are shown as dashed arrows in \cref{fig:multilevel_reasoning}(a).
Among these criteria, static and dynamic stability evaluations with physics reasoning in \cref{eq:stability} are more expensive than evaluating operability.
Thus, we adopt a modified Depth-First Search (DFS) with partial expansion and parallel action mask evaluation. 
At each node (\ie a partial structure), we identify all operable bricks and only sample a subset for parallel stability evaluation. 
If no executable actions are found, we resample until all options are exhausted, then backtrack. 
This partial expansion approach enables efficient discovery of a physically executable assembly sequence $Q$.

% Among these criteria, static and dynamic stability evaluations are more expensive than evaluating manipulability and operability.
% Thus, we use a modified Depth-First Search (DFS) with partial expansion and parallel action mask evaluation. 
% For a brick structure $\hat B_t$ in the disassembly search, we first identify the set of manipulatable and operable bricks ${\hat b_t}$, then randomly sample a subset to evaluate their static and dynamic stability of $\hat B_{t-1}$ and $\hat B_{t}'$ in parallel. 
% If no physically executable actions are found, we resample another subset of bricks for parallel stability evaluation until no manipulatable or operable bricks remain, and then backtrack to its parent node. 
% By partially exploring the pruned action space and parallelizing stability evaluation in the tree search, we can efficiently obtain a physically executable assembly sequence $Q$.
% Then, we incorporate \apex{} \cite{huang2025apexmr} for task, motion, and action reasoning, which are discussed below.

Given an assembly sequence $Q$, we generate a robot executable plan $G$ following the technique presented in \apex{} \cite{huang2025apexmr} as shown in \cref{fig:multilevel_reasoning}(b)-(d).

\myparagraph{Task Reasoning.}
Following the idea in \apex{}, we start from a sequential task plan, in which only one robot is moving at a time, as depicted in \cref{fig:multilevel_reasoning}(b).
% Starting from a sequential plan significantly simplifies the problem complexity and improves collaboration quality.
We distribute the tasks $\{b^a_i, a_i\}$ to different robots by solving an integer-linear program and construct task dependencies, \ie the gray dashed arrow, indicating that a certain skill (pointing to) should be performed after the other (pointing from). 
In addition to manipulation skills discussed in \apex{}, we add perception skills before and after manipulation skills to improve their robustness and detect failures.

\myparagraph{Motion Reasoning.}
Given the sequential task plan, the robot motions (\ie the pose nodes) for each robot to perform the skills are planned using RRT-Connect \cite{RRT-Connect}. 
Since only one robot is moving at a time, the bimanual cooperation is simplified to a single-agent planning problem, which significantly reduces the planning time and improves the quality of the generated collision-free motion plan.

\myparagraph{Action Reasoning.}
Given the collision-free motion plan, we iterate over the pose nodes and construct motion dependencies, \ie the green dashed arrows, by identifying colliding node pairs.
In the end, we construct a TPG with the nodes $V$ and dependency edges $E$, \ie $G=\{V, E\}$, which allows the robots to execute in parallel and collaborate asynchronously.
Each robot only waits when necessary, significantly improving collaboration efficiency while avoiding potential collisions.
Note that, unlike \apex{}, which only considers force feedback, we incorporate perception skills before and after manipulation skills to improve their robustness and detect failures. 
When a failure is detected, the perception skill blocks the TPG execution and keeps detecting until the failure is recovered, \eg addressed by a human operator. 
The system continues, and the TPG execution flow remains intact since all precedence constraints are still satisfied.
% Safety is guaranteed since all robot execution plans and potential collisions are precomputed and avoided.

By integrating physical constraints, \ie dexterity and feasibility, \builder{} generates bimanual robot operations $G$.
Executing $G$ enables the system to efficiently, safely, and robustly construct customized brick assembly structures, bringing virtual brick designs $\hat B$ to real products $B\in\ourspace$.

\section{Experiment}\label{sec:experiment}

\subsection{\pp{} Interface}
\begin{figure*}
    \centering
    \includegraphics[width=\linewidth]{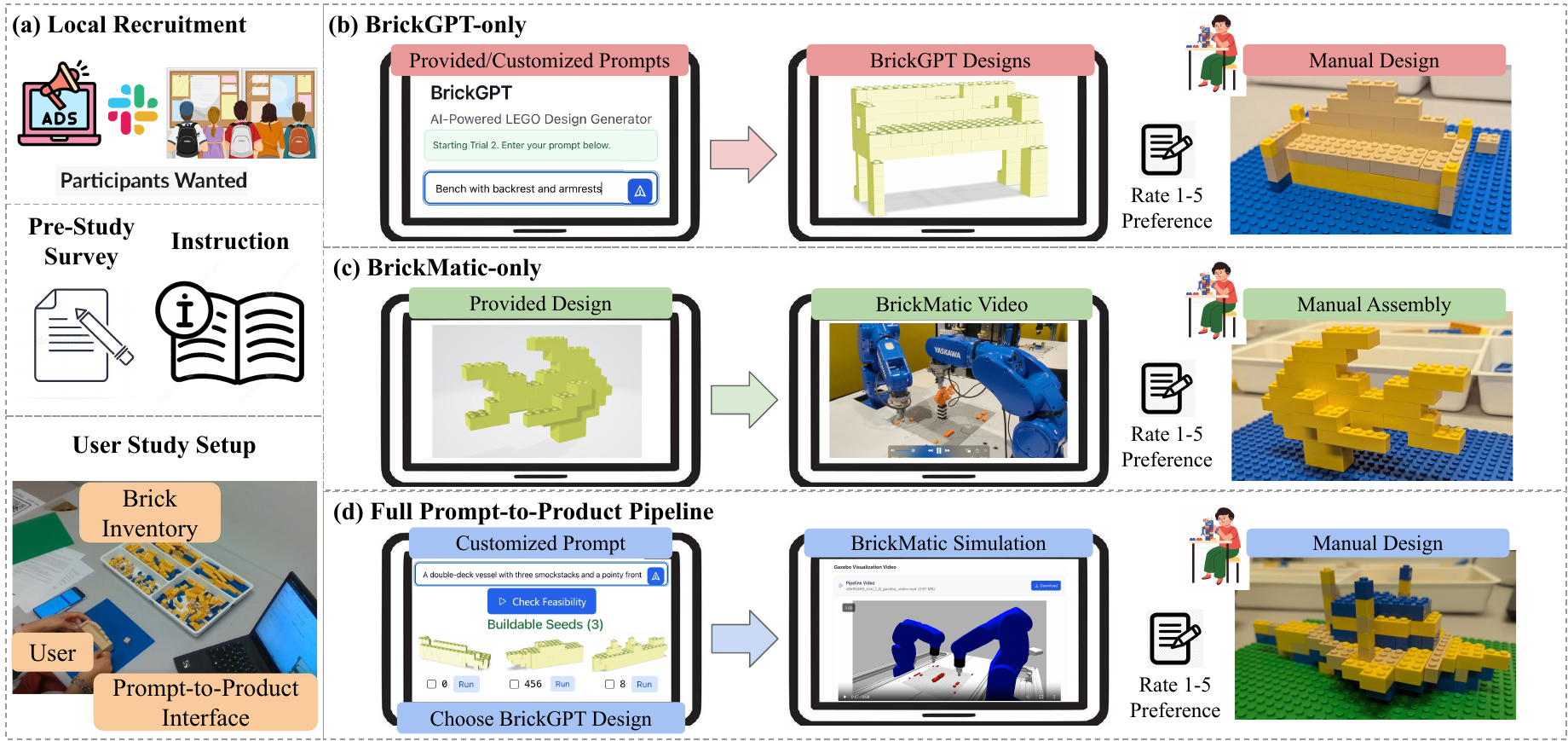}
    \caption{
    \textbf{\pp{} User Study Procedure}. We recruit participants through public online advertisements, internal Slack channels, and physical bulletin boards.
    Each user completes a pre-study survey and interacts with our system. They also fill out a post-study survey in the end.
    % Each user completes a pre-study survey and receives instructions on the study procedure and how to use \pp{}. 
    % In the \LegoGPT{}-only part (b), users enter provided or customized prompts in the \LegoGPT{} web interface and compare the generated design with their manual design. 
    % For the \builder{}-only part (c), users replicate a provided design and compare the pre-recorded robot assembly video with their own manual assembly process. 
    % For the entire \pp{} pipeline, users enter customized prompts, choose from robot-buildable \LegoGPT{}-generated designs, and compare the simulated assembly process with their manual design and assembly. 
    }
    \label{fig:user_study_procedure}
    \vspace{-10pt}
\end{figure*}

The \pp{} is hosted on a server with eight Nvidia RTX A4000 GPUs.
We create a web interface to allow users to interact with \pp{} as shown in \cref{fig:user_study_procedure}(d).
On the front page, users enter the text prompt in the prompt window and \LegoGPT{}++ generates the assembly designs.
Due to the multi-head generation, the system outputs multiple designs and highlights the best one.
Users then select one or more designs based on their preference and proceed with \builder{}.
In our interface, users do not directly interact with the physical \builder{} system.
Instead, we develop a \builder{} digital twin in Gazebo \cite{1389727}, and the web interface displays \builder{} running virtually.
Note that since the digital twin cannot simulate the connections between bricks, we turn off physical interactions between bricks. 
Thus, the virtual \builder{} only uses the manipulation and motion skills, and the execution is deterministic.
After users verify their preferred designs virtually, we run \builder{} offline, which fully utilizes the skill set $\mathcal{A}$, to physically construct the brick assembly products.
Due to the staged architecture of \pp{}, users can also interact with \LegoGPT{} or \builder{} individually.
They can query \LegoGPT{} to generate a design and manually build it as depicted in \cref{fig:user_study_procedure}(b), or they can input a manual brick design and let \builder{} automatically construct it as shown in \cref{fig:user_study_procedure}(c).

% \vspace{-6pt}
\subsection{Module Evaluation}
% \vspace{-2pt}
% We quantitatively evaluate the performance of \LegoGPT{}++ and \builder{} in this section.

\begin{figure}[t]
\begin{minipage}{\linewidth}
        \centering
        \captionof{table}{
        \textbf{Generative Design Comparison.} Results are averaged over 36 open-world unique prompts from the user study in \cref{sec:user_study}. 
        \% Buildable: the percentage of prompts with a generated \builder{}-buildable brick design.
        CLIP: text-image similarity.
        % Time: the time of generating a design given a prompt.
        }
        \label{table:compare_gen}
        \setlength{\tabcolsep}{5pt}
        \resizebox{\linewidth}{!}{
  \begin{tabular}{lcccc}
    \toprule
    Method  & \% Buildable & CLIP \cite{radford2021clip} & Time (s)\\
    \midrule
    \LegoGPT{} \cite{pun2025generating} & 19.4\% & 0.248$\pm$0.036 & \textbf{44.0$\pm$39.0}\\
    \LegoGPT{}++ & \textbf{66.6\%} & \textbf{0.266$\pm$0.029} & 85.7$\pm$48.8\\
  \bottomrule
\end{tabular}
}
    \end{minipage}
    \vspace{-10pt}
\end{figure}

\myparagraph{Generative Assembly Design.}
% We leverage \LegoGPT{} to generate assembly designs from user text prompts.
The multi-head generation in \cref{sec:brickgpt_multihead} improves the quality of design generation.
\Cref{table:compare_gen} shows the comparison between the modified \LegoGPT{}++ and the original \LegoGPT{} \cite{pun2025generating}, which generates one design per prompt.
As shown by the buildable rate, the modified \LegoGPT{}++ has a higher chance of generating \builder{}-buildable designs.
In addition, the CLIP score indicates that the designs from \LegoGPT{}++ better align with the open-world user prompts.
Despite the longer generation time, \LegoGPT{}++ improves the quality of the design generation and enhances the overall \pp{} workflow.

\begin{figure}
\centering
\subfigure[Faucet.]{\includegraphics[width=0.24\linewidth]{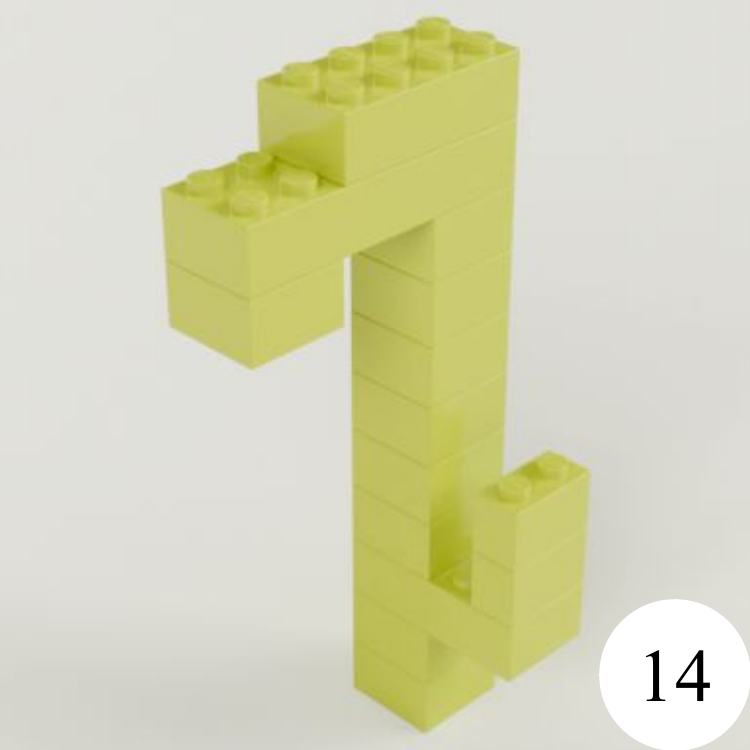}\label{fig:faucet_render}}\hfill
\subfigure[Fish.]{\includegraphics[width=0.24\linewidth]{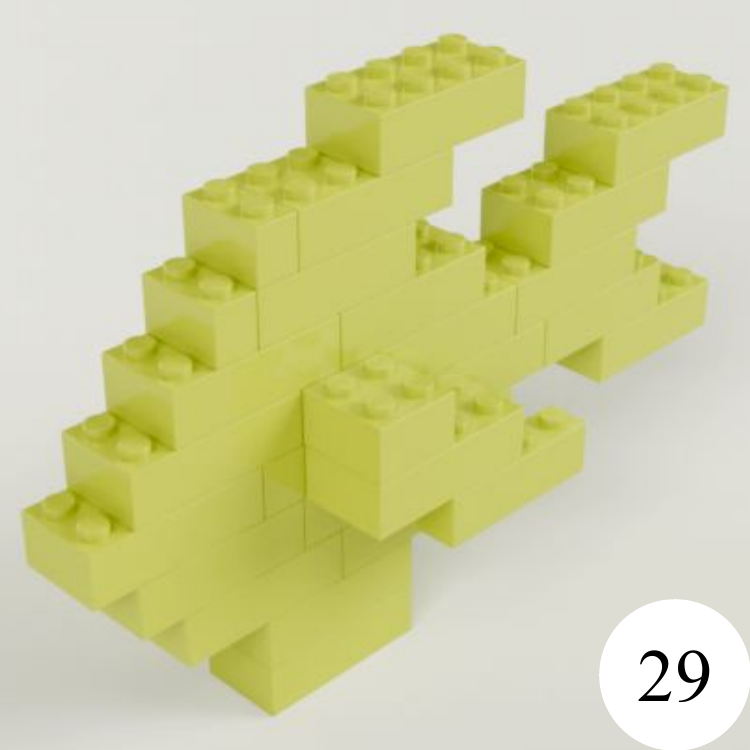}\label{fig:fish_render}}\hfill
\subfigure[Vessel.]{\includegraphics[width=0.24\linewidth]{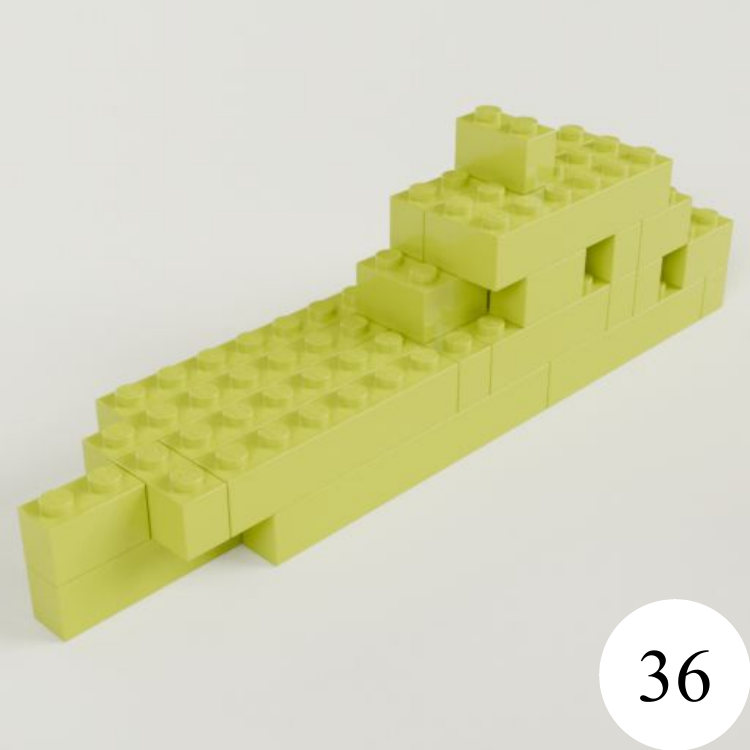}\label{fig:vessel_render}}\hfill
\subfigure[Guitar.]{\includegraphics[width=0.24\linewidth]{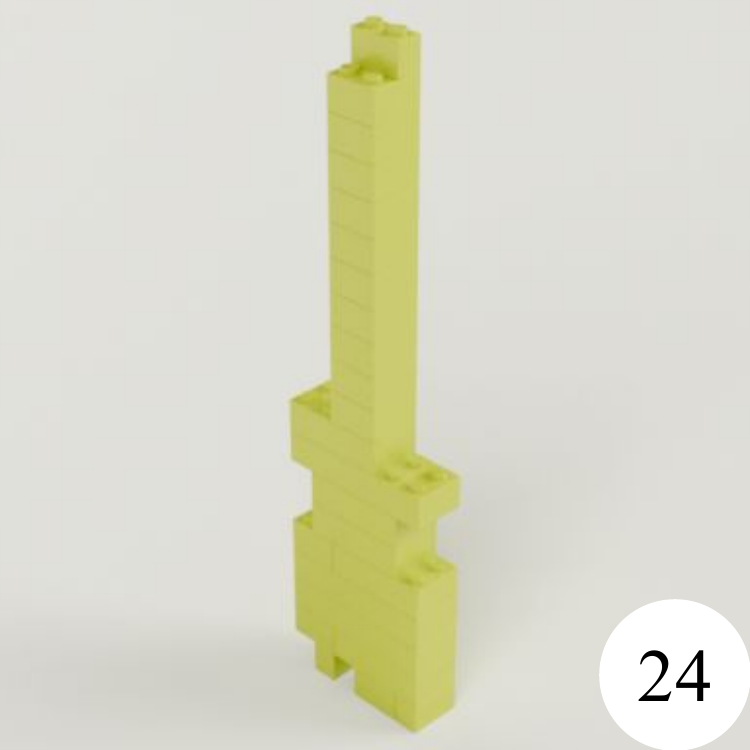}\label{fig:guitar_render}}
\vspace{-5pt}
    \caption{ Brick assembly designs for comparing bimanual assembly construction. The number in each figure indicates the number of bricks required. \label{fig:compare_structures}}
    \vspace{-10pt}
\end{figure}

\myparagraph{Bimanual Assembly Construction}
We compare the presented \builder{} with the dual-arm system in \cite{huang2025apexmr}.
Specifically, \builder{} is equipped with innovative EiF EOATs providing close-proximity visual feedback and has an expanded skill set $\mathcal{A}$ with perception skills for anomaly detection.
\Cref{fig:compare_structures} illustrates brick assembly designs we use to compare the systems.
Here, we define a successful build as a trial in which the system builds the structure without manually stopping or restarting.
As shown in \cref{table:compare_build}, with the integration of perception skills, \builder{} has a significantly higher success rate as it successfully builds all designs in one attempt. 
This is because when a failure happens, \builder{} automatically pauses its construction and actively requests human intervention for failure recovery.
It continues its construction when the failure is addressed without starting over.
On the other hand, the dual-arm system \cite{huang2025apexmr} needs to restart the entire building sequence once a failure happens.
Consequently, it requires significantly more attempts and even fails to successfully build the Fish after many trials.
In addition, the survival length reveals that \builder{} can perform significantly longer horizon assembly tasks without restarting.
Moreover, we compare the planning time for both systems.
Due to the modified DFS in \cref{sec:mr_reasoning}, \builder{} achieves a significantly shorter runtime.
% , outperforming the dual-arm system in \cite{huang2025apexmr}.
% A feasible assembly sequence can be found in 10 and 7 seconds for the fish and vessel (\ie \cref{fig:fish_render,fig:vessel_render}) on an AMD 7840HS laptop CPU. 
% The runtime in task, motion, and action reasoning is the same as \cite{huang2025apexmr}. 

\begin{figure}[t]
\begin{minipage}{\linewidth}
        \centering
        \captionof{table}{ 
        \textbf{Bimanual Construction Comparison.}
        Success Rate: number of trials attempted to have the system successfully build the brick design once without restarting.
        Survival Length: number of bricks (averaged over the attempts) the system assembled without restarting. 
        Time: the time to plan the assembly sequence $Q$, and construct TPG $G$ given the brick design.
        }
        \label{table:compare_build}
        \setlength{\tabcolsep}{5pt}
  \resizebox{\linewidth}{!}{
  \begin{tabular}{lccccc}
    \toprule
    Method & Design & Success Rate & Survival Length & Time (s) \\
    \midrule
    \multirow{4}{*}{Dual-Arm \cite{huang2025apexmr}} & Faucet & 1/5 & 9.2 % (6, 6, 7, 13, 14)
    & 30.0 %4.3+25.7  
    \\
    & Fish & 0/5 & 7.8 % (6, 6, 7, 13, 14)
    & 157.0 %83.5+73.5   
    \\
    & Vessel & 1/3 & 33.7 % (36, 33, 32)
    & 78.4 %26.8+51.6 
    \\
    & Guitar & \textbf{1/1} & \textbf{24} & 50.8 %11.6+39.2 
    \\
    \midrule
    \multirow{4}{*}{\builder{}} & Faucet & \textbf{1/1} & \textbf{14}  & \textbf{27.2} %1.5+25.7 
    \\
    & Fish & \textbf{1/1} & \textbf{29} & \textbf{83.4}%9.9+73.5 
    \\
    & Vessel & \textbf{1/1} & \textbf{36} & \textbf{58.1}% 6.5+51.6
    \\
    & Guitar & \textbf{1/1} & \textbf{24} & \textbf{42.9}%3.7+39.2 
    \\
  \bottomrule
\end{tabular}
}
    \end{minipage}
    \vspace{-10pt}
\end{figure}

% \vspace{-7pt}
\subsection{User Study}
\label{sec:user_study}

We conduct an IRB-approved user study to investigate how the proposed \pp{} can help people create physical assemblies from abstract ideas. 
Specifically, we want to test the following hypothesis: 
\begin{enumerate}
    \item Generative AI (\LegoGPT{}) helps users to bring their abstract ideas into concrete assembly designs. It reduces the required manual effort and expert knowledge.
    
    \item The designs from \LegoGPT{} are good initial designs for users to improve further.

    \item Bimanual robotic assembly (\builder{}) reduces the required manual effort in creating physical products from virtual designs.
    
    \item Users would prefer to build by themselves for fun if they were only to build a few assemblies. 
    They would prefer robotic construction for mass production.
\end{enumerate}

\myparagraph{Tasks.}
Participants were divided into three groups: \LegoGPT{}-only, \builder{}-only, and the full \pp{}. In the \LegoGPT{}-only group, users submitted a prompt, created a manual design, and compared it with the best generated result (\cref{fig:user_study_procedure}(b)). In the \builder{}-only group, users assembled a given design with 3D visualization support (\cref{fig:user_study_procedure}(c)), then watched a video of the \builder{} performing the same task and rated their preference. In the full \pp{} group, users entered a prompt, created a manual design, selected a generated design, and compared the manual design with the one assembled in simulation by \builder{} (\cref{fig:user_study_procedure}(d)).

\myparagraph{Participants.}
We recruited 21 participants (ages 18–31; 6 female and 15 male), primarily with technical backgrounds, while some came from the arts or architecture fields. 
LEGO experience ranged from none (2 participants) to advanced (4 had built sets >1000 pieces). 
In our study, 8 users tried \LegoGPT{}, 7 used \builder{}, and 6 interacted with the full pipeline.

\myparagraph{Procedure.}
Each user interacted with our system for $\sim$30 minutes. 
For \LegoGPT{}-only, each user had 4 design attempts, including 2 provided prompts and 2 customized prompts in one of ten categories (Bookshelf, Car, Chair, Guitar, Sofa, Table, Vessel, Bench, Bottle, Bus). 
For \builder{}-only, each user assembled 2-4 brick designs, depending on the time needed to replicate the exact design.
For the entire \pp{} pipeline, each user tried two customized prompts from two categories (Guitar, Vessel).
All participants filled out a post-study survey, rating the system usability and favorability.

% Lastly, they answer a series of post-study questions on the overall quality, usability, and favorability of our system.

\begin{figure*}
    \centering
    \includegraphics[width=\linewidth]{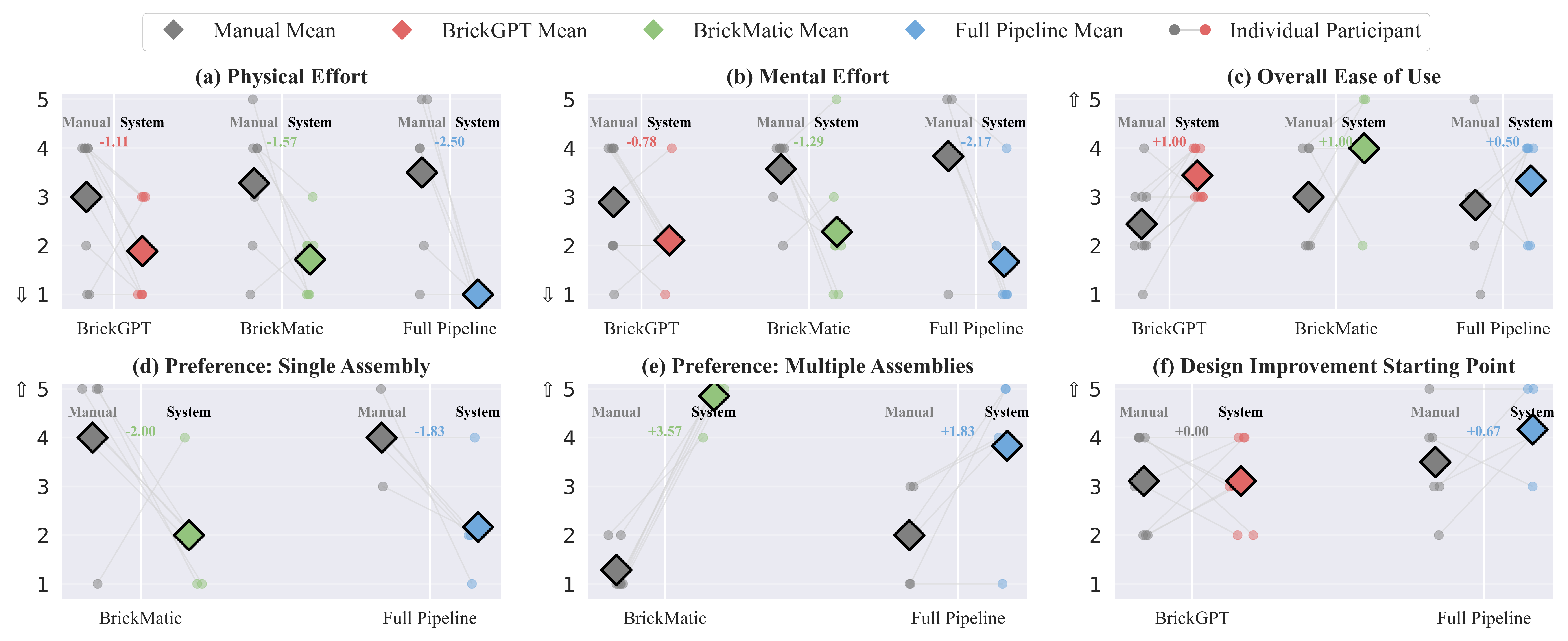}
    \caption{ \textbf{\pp{} User Study Results.} Individual and mean ratings of key survey questions for BrickGPT-only, BrickMatic-only, and entire Prompt-to-Product. (a) Physical demands of designing or building brick assembly structures. (b) Mental demands of designing or building brick assembly structures. (c) The overall ease of use. (d) User's preference when only building one brick design. (e) User's preference when building multiple designs. (f) User's rating of how well the manual or \LegoGPT{}-generated design serves as a starting point for improvement.}
    \label{fig:user-study-result}
    \vspace{-10pt}
\end{figure*}

\myparagraph{Results.}
User responses to key survey questions are summarized in \cref{fig:user-study-result}. Overall, the \pp{} system is well-received and notably reduces user effort, especially when assembling multiple designs.
For Hypotheses 1 and 3, Wilcoxon signed-rank tests on 5-point Likert responses show significant reductions in both physical and mental effort when using \pp{} compared to manual creation. \LegoGPT{} significantly reduces physical (p = 0.037) and mental effort (p = 0.041). \builder{} yields a significant reduction in physical effort (p = 0.023) and a marginal reduction in mental effort (p = 0.055). The full pipeline shows the strongest effects, with highly significant reductions in both physical and mental effort (p = 0.021 for both).
For Hypothesis 4, users prefer using \builder{} (p = 0.008) or the full pipeline (p = 0.021) when building multiple structures, but slightly prefer manual assembly for a single design with \builder{} (p = 0.109) or the full pipeline (p = 0.039).
For Hypothesis 2, no significant preference is observed between manual and \LegoGPT{}-generated designs as starting points (p = 0.5). However, with the full pipeline, users show a slight preference for generated designs (p = 0.097), suggesting they are at least equally suitable—or potentially better—for refinement.

Manual creation time ranged from under 2 to over 7 minutes, reflecting differences in user experience with brick assembly. 
Users generally preferred their manual designs over generated ones for custom prompts (3.79 vs. 2.86), while slightly preferred generated designs for provided prompts (3.44 vs. 3.50). 
Custom prompts were often highly creative (\eg a five-wheeled car, a tree-shaped bookshelf), challenging the system.
Notably, all users expressed strong interest in using the \pp{} system again.
% More examples of user study prompts, results, and videos are available at \url{https://prompt2product.github.io}.

\section{Discussion}\label{sec:discussion}
We propose \pp{}, a pipeline that translates abstract text prompts into physical brick assemblies, reducing required manual effort and expert knowledge. 
While promising, \pp{} has limitations to be address in future work.

\myparagraph{Generative Design:}
\pp{} currently supports only brick assemblies. 
While users can modify the brick inventory $\mathcal{I}$, non-brick components are not supported, limiting the system from generating more vivid designs. 
Additionally, it is restricted to categories in \SLV{}, leading to degraded performance on open-ended prompts—a limitation frequently noted by users during the study. 
Hence, we will explore more diverse 3D datasets and expand the generative capabilities to include varied components and higher-fidelity assemblies that better match user intent.

\myparagraph{System Dexterity:}
\pp{} showcases state-of-the-art capabilities in long-horizon and dexterous manipulation tasks, but still falls short when compared to human-level dexterity, \ie user-constructable structures remain unbuildable by the current system (\cref{table:compare_gen}).
Thus, we will enhance \builder{} by incorporating additional skills, such as in-hand assembly, failure recovery, and coordinated subassembly.

\section*{Acknowledgments}
The authors would like to thank Ken Goldberg, Oliver Kroemer, and Jean Oh for their discussions.  
This work is in part supported by the Manufacturing Futures Institute, Carnegie Mellon University, through a grant from the Richard King Mellon Foundation.

% \clearpage
% \newpage
% Bibliography
{
    \small
    \bibliographystyle{plainnat}
    \bibliography{ref}
}

% Uncomment for the final version
% \newpage

% \section{Biography Section}
% If you have an EPS/PDF photo (graphicx package needed), extra braces are
%  needed around the contents of the optional argument to biography to prevent
%  the LaTeX parser from getting confused when it sees the complicated
%  $\backslash${\tt{includegraphics}} command within an optional argument. (You can create
%  your own custom macro containing the $\backslash${\tt{includegraphics}} command to make things
%  simpler here.)
 
% \vspace{11pt}

% \bf{If you include a photo:}\vspace{-33pt}
% \begin{IEEEbiography}[{\includegraphics[width=1in,height=1.25in,clip,keepaspectratio]{fig1}}]{Michael Shell}
% Use $\backslash${\tt{begin\{IEEEbiography\}}} and then for the 1st argument use $\backslash${\tt{includegraphics}} to declare and link the author photo.
% Use the author name as the 3rd argument followed by the biography text.
% \end{IEEEbiography}

% \vspace{11pt}

% \bf{If you will not include a photo:}\vspace{-33pt}
% \begin{IEEEbiographynophoto}{John Doe}
% Use $\backslash${\tt{begin\{IEEEbiographynophoto\}}} and the author name as the argument followed by the biography text.
% \end{IEEEbiographynophoto}

% \vfill

\end{document}